\documentclass[11pt,DIV=14,parskip=half,a4paper]{scrartcl}

\newcommand{\redactionmode}{1}

\usepackage[utf8]{inputenc}
\usepackage[textsize=tiny,obeyFinal]{todonotes}
\usepackage[backend=biber,style=numeric,sorting=none,isbn=false,url=false,eprint=false,citestyle=numeric-comp,backref=false,giveninits]{biblatex}

\usepackage[affil-it]{authblk} 
\usepackage{amsmath}
\usepackage{amsthm}
\usepackage{amssymb}
\usepackage{wasysym}
\usepackage{textcomp,gensymb}  
\usepackage{mathtools}
\usepackage{nicefrac}
\usepackage{csquotes}

\usepackage{caption}
\usepackage{subcaption}
\usepackage{xifthen}
\usepackage{tikz}
\usepackage{multicol}
\usepackage[outline]{contour}
\usepackage{pgfplots}
\usepackage[super]{nth}
\usepackage[noend,boxed]{algorithm2e}
\usepackage{multirow}
\usepackage{numprint}
\usepackage{ifthen}
\usepackage{hyperref}
\usepackage{cleveref}

\usetikzlibrary{fit}
\usetikzlibrary{matrix}
\usetikzlibrary{shapes}
\usetikzlibrary{positioning}
\pgfplotsset{compat=1.17}

\DeclareMathOperator{\Ima}{Im}

\Crefname{algocf}{Algorithm}{Algorithms}
\Crefname{equation}{Eq.}{Eq.}

\npthousandsep{,}

\newcommand{\generictodo}[3]{\todo[tickmarkheight=0.15cm,shadow,inlinewidth=0.97\textwidth,color=#1,#2]{#3}}

\newcommand{\TODO}[2][Who?]{\ifthenelse{\isempty{#2}}{\smallskip\generictodo{red!90}{inline, inlinewidth=0.14\textwidth, caption={Content Missing}}{\color{white}\large{TODO}}\smallskip}{\generictodo{orange!80}{}{\textbf{\underline{TODO} (#1)}\\[0.5em]#2}}}

\newcommand{\ite}{i.\,e.\xspace}
\newcommand{\eg}{e.\,g.\xspace}
\newcommand{\cf}{cf.\xspace}

\newcommand{\Prob}[2][]{\operatorname{P}\ifthenelse{\isempty{#2}}{}{\left(#2\ifthenelse{\isempty{#1}}{}{\;\middle|\;#1}\right)}}
\newcommand{\Freq}[2][]{\operatorname{f}\ifthenelse{\isempty{#2}}{}{\left(#2\ifthenelse{\isempty{#1}}{}{\;\middle|\;#1}\right)}}

\newcommand{\featspace}{\ensuremath{\mathfrak{F}}\xspace}
\newcommand{\group}{\ensuremath{\Pi}\xspace}

\newcommand{\errorlimit}{\ensuremath{\varepsilon}\xspace}

\newcommand{\dist}{\ensuremath{\Psi}\xspace}

\newcommand{\trans}{\ensuremath{T}\xspace}

\newcommand{\R}{\mathbb{R}\xspace}
\newcommand{\N}{\mathbb{N}\xspace}
\newcommand{\congraph}{\ensuremath{\mathcal{G}\xspace}}

\makeatletter
\newcommand{\@minipagerestore}{\setlength{\parskip}{\medskipamount}}
\makeatother

\newcommand{\redact}[2][]{%
    \ifcase\redactionmode%
        \ifthenelse{\isempty{#1}}%
            {%
              {\footnotesize\color{black!40}#2}%
            }%
            {%
                \ifvmode%
                  \begin{minipage}{0.3\textwidth}#1\end{minipage}%
                  \hfill%
                  \begin{minipage}{0.6\textwidth}\footnotesize\color{black!40}#2\end{minipage}%
                \else%
                  \par%
                  \begin{minipage}{0.3\textwidth}#1\end{minipage}%
                  \hfill%
                  \begin{minipage}{0.6\textwidth}\footnotesize\color{black!40}#2\end{minipage}%
                  \par%
                \fi
            }
    \or%
        #1%
    \or%
        #2%
    \fi%
}

\tikzstyle{state} = [rectangle, rounded corners, minimum height=1cm,text centered, draw=black, fill=white, text width=2.8cm]

\graphicspath{{images/}}

\bibliography{references}

\title{Unsupervised Learning of Invariance Transformations}

\author[1]{Aleksandar~Vu\v{c}kovi\'c}
\author[2]{Benedikt~Stock}
\author[1]{Alexander~V.~Hopp}
\author[1]{Mathias~Winkel}
\author[3,*]{Helmut~Linde}

\affil[1]{Merck~KGaA, Darmstadt, Germany}
\affil[2]{University~of~Oxford, Oxford, United~Kingdom}
\affil[3]{Transylvanian~Institute~of~Neuroscience, Cluj-Napoca, Romania}
\affil[*]{\emph{Corresponding Author}}

\date{\today}

\begin{document}

\maketitle

\begin{abstract}
    The need for large amounts of training data in modern machine learning is one of the biggest challenges of the field.
    Compared to the brain, current artificial algorithms are much less capable of learning invariance transformations and employing them to extrapolate knowledge from small sample sets. 
    
    It has recently been proposed that the brain might encode perceptual invariances as approximate graph symmetries in the network of synaptic connections. Such symmetries may arise naturally through a biologically plausible process of unsupervised Hebbian learning.

    In the present paper, we illustrate this proposal on numerical examples, showing that invariance transformations can indeed be recovered from the structure of recurrent synaptic connections which form within a layer of feature detector neurons via a simple Hebbian learning rule. 
    
    In order to numerically recover the invariance transformations from the resulting recurrent network, we develop a general algorithmic framework for finding approximate graph automorphisms. We discuss how this framework can be used to find approximate automorphisms in weighted graphs in general.
\end{abstract}

\section{Introduction}

Many cognitive tasks require an agent to establish a model of its surroundings based on sensory inputs.
This involves the reconstruction of semantic and functional relationships inherent in the environment from data generated by perceptual processes which are not fully structure-preserving.
Based on the assumption that several independent entities contribute to and interact in generating the percepts, it is a promising strategy to decorrelate this data and thus reveal some of the structure present in the environment.

An approach often followed in practice is to consider the observed data points as vectors in a suitable vector space and find a convenient base in which to represent them.
Well-known algorithms based on this idea are, for example, Principal Component Analysis \cite{Bro:2014,Jolliffe:2005,Jolliffe:2016}, Independent Component Analysis \cite{Lewicki:2000,Hyvarinen:2000}, Non-negative Matrix Factorization \cite{lee_learning_1999,WeixiangLiu:2003,Hoyer:2002} and several variants of Sparse Coding \cite{Elad:2010,Hoyer:2002,Lee:2007}.
The basis vectors found by such algorithms are typically called \emph{atoms} and the collection of atoms is called \emph{dictionary}, a nomenclature we will adhere to in the following.

When applied to natural data, these algorithms typically find atoms of significantly lower complexity than the entities they try to reconstruct.
Even though there is, in fact, still a rich structure of correlations present after transforming to a basis of such simple atoms \cite{geisler_visual_2008,Gerhard:2013}, today's algorithms are not able to fully exploit this implicit information in the construction of more complex entity representations.

The apparent difficulty in reconstructing complex entities from unstructured data stems to a large extent from the combinatorial explosion in the number of possibilities for such an entity to appear.
The appealing approach of extending the dictionary to contain more complex atoms is impractical due to limited computing power and memory capacity and due to the much more fundamental problem that suitable atoms of a certain complexity do not appear sufficiently often in the data to be identified.

A promising approach is to model invariances present in the data explicitly.
In practice, such invariances may include transformations like rotations or translations. 
Their explicit inclusion in the model ensures that observations that only differ by a transformation of this kind can still contribute to the formation of the same atoms and to discovering structure in new transformed observations.
Examples of such models are Transform-invariant NMF \cite{tnmf_github}, Transform-invariant Restricted Boltzmann Machines \cite{Sohn2012LearningIR}, or the generalized pooling mechanism presented in \cite{anselmi_unsupervised_2016}.
Yet, for such algorithms to work, the invariance transformations must be known explicitly in the first place.
While this is the case for basic geometric transformations, it is desirable to have a general way of learning transformations from the data that does not require a-priori knowledge.

Several methods have been proposed to learn invariance transformations directly from data.
Most of them rely on observing at least two versions of the same entity.
Many of these approaches are based on the assumption of time continuity where two subsequent frames of data are likely to show the same objects. Infinitesimal transformations between images can be learned, \eg, via a temporal Hebbian learning rule \cite{foldiak_learning_1991}, by analyzing the correlation structure between features in subsequent frames \cite{bethge_unsupervised_2007,memisevic_learning_2013}, or by fitting an explicitly time-dependent sparse multi-layer model \cite{cadieu_learning_2012}.
Other studies approached the problem by learning the generators of a Lie group to best approximate observations which differ by transformations chosen from the group \cite{Rao_Learning_1998, miao_learning_2007, cohen_learning_2014}.

Other approaches are contrastive learning, which attempts to model similarity or dissimilarity relations between data points \cite{le-khac_contrastive_2020}, and labeled graph matching \cite{von_der_malsburg_pattern_statistical_1986, von_der_malsburg_pattern_1988, Westphal_feature-driven_2008, lades_distortion_1993}, which is a method for image recognition where objects are represented as sets of features with distance relations between them; features are represented by nodes and relations by edges, and some transformations can be efficiently dealt with in graph view.

Apparently, the brain implements a much more effective approach to handling invariance transformations, given that humans can recognize objects or sounds under a wide variety of perceptual alterations and based on only a few training samples.
While the algorithms behind this ability are unknown to science, it is clear that the brain differs structurally in significant ways from today's artificial neural networks.
In particular, the brain relies heavily on recurrent neural connections: Even in the visual cortex, synapses from the lateral geniculate nucleus of the thalamus (i.e., the feed-forward connections) form only 5–10\% of the excitatory synapses \cite{douglas_recurrent_2007}.
The overwhelming majority of neocortical synapses connect neurons with neighbouring excitatory neurons located in the same cortical area and thus constitute recurrent connections. 

In a recent publication, it was proposed that unsupervised Hebbian learning might suffice to encode invariance transformations in this recurrent cortical network \cite{Linde:2021}. 
The central idea is that each invariance transformation is reflected in a graph symmetry of the network: An invariance transformation acting on the space of possible observations leaves the probability distribution from which samples are drawn unaltered. The correlations between features of these observations are effectively two-dimensional projections of this probability distribution and thus they are also invariant under the transformation. Hebbian learning on the synapses between feature detectors turns these correlations into a weighted graph of synaptic strengths, which is then also invariant under the transformation. The problem of finding invariance transformations based on a set of observations is thus reduced to finding (approximate) symmetries in the graph of recurrent synaptic strengths. 

The scope of the present paper is to substantiate this theoretical concept by testing it on numerical examples.
We construct a simple recurrent network with synthetic data and show that invariances in the training data can be retrieved from the graph of synaptic connections.
While our transformation-finding algorithm is very different from how the brain works, it illustrates that the information about invariances is encoded in the symmetries of the graph.
This encoding does in fact appear to be biologically plausible and the reader is referred to \cite{Linde:2021} for a discussion of related experimental observations from neuroscience.
We emphasize that our study aims at demonstrating the feasibility of the described encoding scheme for invariances, but it cannot answer the question how this encoding can be used efficiently by the brain or by artificial algorithms to solve practically relevant computational problems.

This paper is organized as follows: We elaborate on the concept of invariance transformations in \Cref{sec:transformations} and introduce the algorithmic framework for finding the invariances in \Cref{sec:Method}.
In \Cref{sec:Numerical} we present the results of our numerical study which constitute the first computational corroboration of the concept proposed in \cite{Linde:2021}.
Pseudocode of the discussed algorithms can be found in \Cref{sec:Algorithms}.

\section{Invariance Transformations} \label{sec:transformations}
We begin by defining the term \emph{invariance transformation} following the assumptions made in~\cite{Linde:2021}.

We consider a learning process in which observations are repeatedly sampled from a discrete probability distribution $\dist: \featspace \to[0,1]$ defined on a \emph{feature space} $\featspace = \{0,1\}^n$ for some $n \in \N$ and $n \gg 1$.
In our discussions below, we interpret each dimension of $\featspace$ as a pixel in an image which can be off or on.
Yet, as the name \enquote{feature space} suggests, the proposed concept is also applicable to the case where each dimension represents aggregated information like features in an image that are either present or not and can also be generalized to other perceptual modalities.

We define an invariance transformation as a permutation of the coordinates of $\featspace$ which leaves~$\dist$ unaltered.
More precisely, an invariance transformation is a permutation $\trans:\featspace\to\featspace$ such that $\dist(Tx)=\dist(x)$ for all $x\in \featspace$.

This implies that the consecutive application of two or more transformations is also a valid transformation.
As the inverse of a transformation is also a transformation, the set of transformations forms a subgroup of the symmetric group on~$n$ elements.
Thus, the set of invariance transformations is finite and it is sufficient to find a set of generators of this subgroup.

Our objective is to learn invariance transformations based on a limited number of samples drawn from the probability distribution $\dist$.
Unfortunately, it is not possible to measure $\dist$ exactly as it is only possible to estimate it based on the samples that we draw from it.
An approach to dealing with this problem is based on the observation that symmetries of $\dist$ are \enquote{inherited} by its projections to lower-dimensional spaces.
In particular, consider the one- and two-dimensional marginal distributions
\begin{equation} \label{eq:marginal_def}
\dist_{i}(x_i) =\mkern-8mu\sum_{x_j\in\mathbb{Z}:j\neq i}\mkern-8mu\dist(x_1,\dots,x_n)
    \quad \text{and} \quad
\dist_{i,j}(x_i,x_j) =\mkern-8mu\sum_{x_k\in\mathbb{Z}:k\neq i,j}\mkern-8mu\dist(x_1,\dots,x_n)
\end{equation}
for $i,j\in\{1,\dots,n\}, i\neq j$.
Then, if $T$ is an invariance transformation of $\dist$ acting as
\begin{equation}
    T(x_1,\dots,x_n) = (x_{\tau(1)},\dots,x_{\tau(n)})
\end{equation}
for some permutation $\tau$ of $\{1,\dots,n\}$, we have 
\begin{equation} \label{equ:marginal_sym}\\
    \begin{split}
        \dist_{i}(y) &= \dist_{\tau(i)}(y) \quad \text{for} \quad y \in \{0,1\} \quad \text{and} \\
        \dist_{i,j}(y) &= \dist_{\tau(i),\tau(j)}(y) \quad \text{for} \quad y \in \{0,1\}^2.    
    \end{split}
\end{equation}

The proof of these statements is straight-forward and can be found in the appendix of \cite{Linde:2021}. 
Since the marginal distributions $\Psi_{i,j}$ are defined on only four points each, a limited number of samples can suffice to estimate them with good accuracy.

Moreover, for fixed indices $i,j$, the distribution $\dist_{i,j}$ is uniquely characterized by $\dist_{i,j}(1,1)$, $\dist_{i}(1)$ and $\dist_{j}(1)$.

Our approach is to find approximate invariance transformations of these two-dimensional marginal distributions.

We now argue that approximations of symmetries of these marginal distribution can be found by calculating approximate automorphisms of a suitable node- and edge-weighted undirected graph~$\congraph=(V,E)$ with weighted adjacency matrix~$A$.
As the graphs considered in this paper will not have loops, we slightly misuse notation by interpreting the node weights as edges from a node to itself and hence via the corresponding diagonal entries of $A$.
Using this notation, an automorphism of such a graph $\congraph$ is a permutation $\pi:V\to V$ of the nodes such that \begin{enumerate}
    \item $(i,j)\in E$ if and only if $(\pi(i),\pi(j))\in E$,
    \item $A_{i,j}=A_{\pi(i),\pi(j)}=A_{\pi(j),\pi(i)}$ for all $\{i,j\}\in E$, and
    \item $A_{i,i}=A_{\pi(i),\pi(i)}$ for all $i\in V$.
\end{enumerate}

Note that the problem of finding graph automorphisms is computationally equivalent to the closely related graph isomorphism problem, and that their exact complexity status is not known~\cite{Kobler:1992,Kobler:1993,Arvind:2006}.
In particular, no polynomial-time algorithm for solving either of the problems is known.

In order to translate our problem into the language of graphs, we define the \emph{concurrence graph} $\congraph$ as the complete graph on the set $V=\{1,\dots,n\}$ with weighted adjacency matrix $A_{i,j}\coloneqq\dist_{i,j}(1,1)$ for $i,j\in V, i\neq j$ and node weights $A_{i,i}\coloneqq\dist_i(1)$ for $i\in V$.

From this definition, it follows that each automorphism of the concurrence graph yields an invariance transformation of the two-dimensional and one-dimensional marginal distribution and vice versa.
Thus, by approximating this graph from observed samples of $\dist$ and identifying approximate symmetries of its two-dimensional marginal distribution, we can find candidates for potential invariance transformations.
Note however, that the inverse is not always true:
Observing a symmetry of the marginal distribution does not imply the existence of a corresponding symmetry in $\dist$.
We give an example for this phenomenon in \Cref{sec:GeneralExperimentalSetup}.

\begin{figure}[tb]
    \hfill
        \begin{subfigure}[t]{0.48\textwidth}
        \centering
        \includegraphics[width=0.9\textwidth]{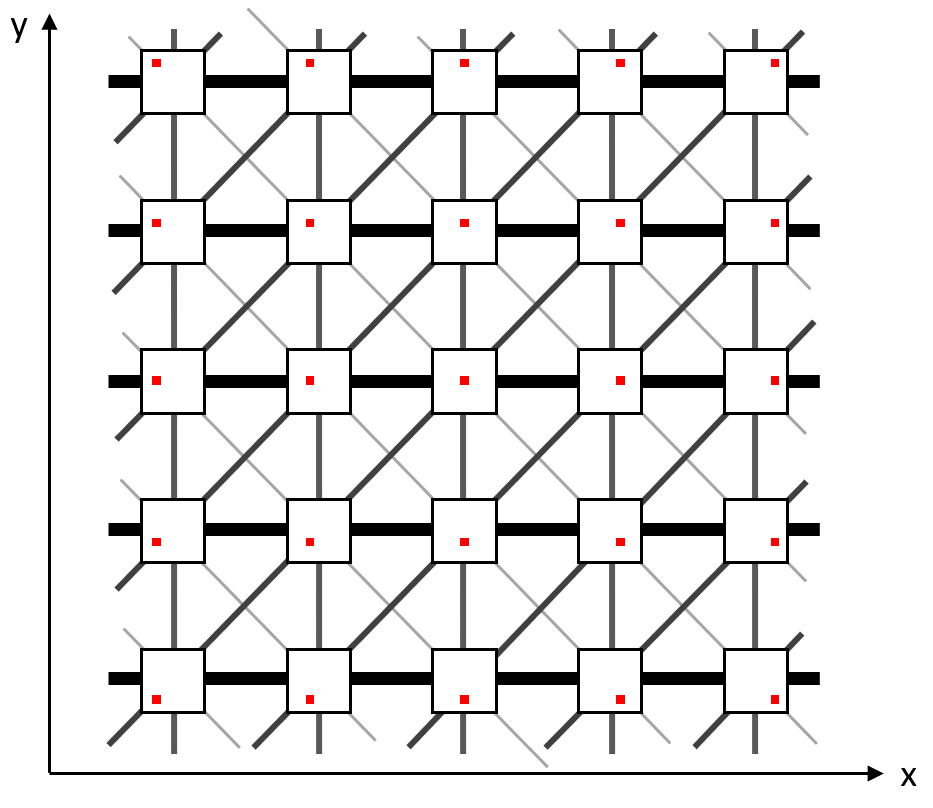}
        \caption{}
        \label{fig:example:1}
    \end{subfigure}
    \hfill
    \begin{subfigure}[t]{0.48\textwidth}
        \centering
        \includegraphics[width=0.9\textwidth]{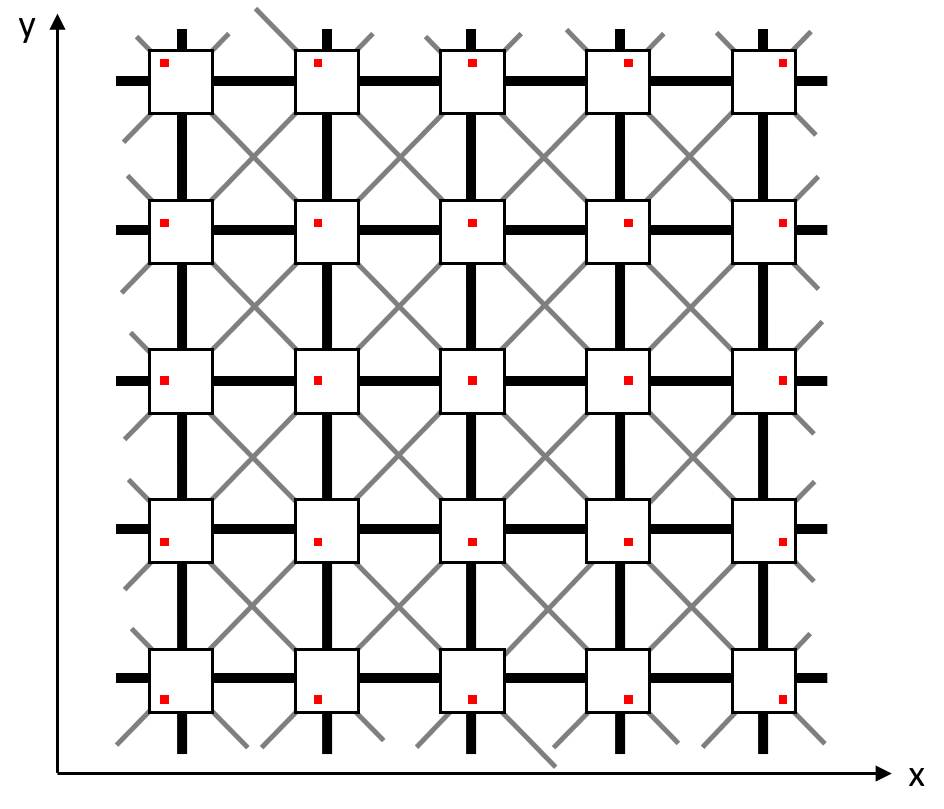}
        \caption{}
        \label{fig:example:2}
    \end{subfigure}
    \caption{Two schematic examples of concurrence graphs. The feature space $\featspace$ in each example consists of pixels at different positions $(x,y)$ in an image.  Each little box represents a feature detector which responds to a single pixel and the position of the red dot within the box corresponds to the position of this pixel in the image plane. The line strength between boxes represents the correlation between the corresponding features and thus also the weight of an edge in the concurrence graph. Correlations are typically strongest between neighboring pixels and therefore lines between distant boxes are omitted for better readability. }
    \label{fig:graph}
\end{figure}

\Cref{fig:graph} shows a simple example of how certain invariances give rise to corresponding graph symmetries. The graph in \Cref{fig:example:1} exhibits translational symmetries but no rotational one. 
It might have been trained, for example, on images containing upright letters at random positions. 
The fact that connections between horizontal neighbors are stronger than those between vertical ones is then a direct reflection of the internal correlation structure of the letters used. For example, the letter \enquote{E} contains more horizontal line segments then vertical ones and thus induced stronger correlations between horizontal neighbors.
The graph in \Cref{fig:example:2} is symmetric under translations as well as rotation by $90\degree$ and reflections.
It might have been trained on images of symmetric letters at random positions and random orientations in $90\degree$ steps, for example.

\section{Algorithmic Framework} \label{sec:Method}

\subsection{Mathematical Background}
Surprisingly, there are only few other approaches for calculating approximate automorphisms of graphs.
The vast majority of the research in this field of graph theory is concerned with either the related \emph{graph isomorphism} problem or with finding \emph{exact} automorphisms.
Although there are results related to finding and repairing imperfect automorphisms by adding additional edges~\cite{Leifer:2021}, by defining symmetry in graphs using other concepts than automorphisms \cite{Morone:2019,Boldi:2021}, or by applying techniques from the field of fractional graph theory \cite{Kersting:2014,Takapoui:2016,Scheinerman:2008,Grohe:2021}, most techniques developed for the graph isomorphism problem cannot be applied for the graph automorphism problem.
For more details on different techniques used for the exact and inexact graph isomorphism problem, we refer to \cite{Shervashidze:2011,McKay:2014,Abu-Aisheh:2015,Morone:2019} and the references therein.

Our symmetry calculation algorithm employs \emph{(mixed) integer programming}, a widely used tool in the area of discrete optimization \cite{Schrijver:2011,Conforti:2014}.

Our formulation is based on an integer program presented in \cite{Takapoui:2016} that allows for calculating exact graph automorphisms of a graph with adjacency matrix $A$ by solving the feasibility problem \Cref{equ:APeqPA}.

\begin{minipage}{0.45\textwidth}
    \begin{equation} 
    \label{equ:APeqPA}
    \begin{array}{lr@{\hspace*{0.5em}}ll}
        \text{find}  & P &\\
        \text{subject to}   & 0             &=AP-PA \\
                            & P\mathbf{1}   &=\mathbf{1} \\
                            & P^T\mathbf{1} &=\mathbf{1} \\
                            & P             &\in\{0,1\}^{n\times n}
    \end{array}
    \end{equation}
\end{minipage}
\hfill
\begin{minipage}{0.45\textwidth}
    \begin{equation}
    \label{equ:Exact_Integer_Program}
    \begin{array}{lr@{\hspace*{0.1em}}ll}
        \text{min}  & z &\\
        \text{subject to}   & z             &\geq ||AP-PA||_{\infty} \\
                            & P\mathbf{1}   &=\mathbf{1} \\
                            & P^T\mathbf{1} &=\mathbf{1} \\
                            & P             &\in\{0,1\}^{n\times n}
    \end{array}
    \end{equation}
\end{minipage}

Here, $\mathbf{1}$ denotes the vector with each entry being equal to $1$ of suitable dimension.
In this formulation, a matrix $P\in\{0,1\}^n$ is calculated subject to three conditions.
The conditions $P\mathbf{1}=\mathbf{1}$ and $P^T\mathbf{1}=\mathbf{1}$ ensure that $P$ is a permutation matrix, while the condition $0=AP-PA$ guarantees that $P$ is in fact an automorphism of the graph.

Consequently, this problem has a solution if and only if $P$ encodes an automorphism of a graph with adjacency matrix $A$ \cite{Takapoui:2016}.

However, we are not interested in finding exact, but approximate automorphisms.

To this end, we relax the constraint $0=AP-PA$ by minimizing the maximum norm of its right-hand side, which we achieve by bounding it by the continuous variable $z$ and minimizing that instead.

This yields the mixed integer program (MIP) \Cref{equ:Exact_Integer_Program}, where $||\cdot||_\infty$ denotes the maximum norm.

Note that the non-linear expression in \Cref{equ:Exact_Integer_Program} can be reformulated into $2n^2$ linear expressions. %

The corresponding integer program thus has $n^2+1$ variables (each entry of $P$ as well as $z$) and $2n^2+2n$ constraints (bounding the value of $z$ and forcing the row and column sums of $P$ to be $1$).
Further note that this formulation is not only applicable for concurrence graphs as defined in \Cref{sec:transformations} but general node- and edge-weighted undirected graphs.

\subsection{Solving the MIP in Practice} \label{sec:SolveMIPInPractice}

For the remainder of this paper, we consider the special case of concurrence graphs.
Thus, we let $\congraph$ denote a concurrence graph and let $A$ denote its weighted adjacency matrix with node weights on the diagonal.

For such a graph, solving \Cref{equ:Exact_Integer_Program} in practice turns out to be hard as the matrix $A$ is typically dense and all $n!$ permutation matrices $P\in\R^{n\times n}$ are feasible solutions, \ite solutions that fulfill all constraints.
To overcome this difficulty, we implement an additional pre-processing routine.
This routine constructs an incomplete permutation $\tilde{P}$ by trying to fix the image of as many nodes of $\congraph$ under $\tilde{P}$ as possible.
Formally, such an incomplete permutation is an injective function $\tilde{P}:\tilde{V}\to V$ defined on a subset $\tilde{V}\subseteq V$.
This guarantees that it is possible to extend an incomplete permutation $\tilde{P}$ to a permutation on $V$.

This enables us to consider a reduced variant of the MIP in \Cref{equ:Exact_Integer_Program} that only needs to map nodes that are not already mapped by $\tilde{P}$ onto nodes that are not already in the image of $\tilde{P}$.
More precisely, given an incomplete permutation $\tilde{P}_1:\tilde{V}_1\to V$, the MIP has to find another incomplete permutation $\tilde{P}_2:\tilde{V}_2\coloneqq V\setminus \tilde{V}_1\to V$ such that the function $P:V\to V$ that acts as $\tilde{P}_1$ on $\tilde{V}_1$ and as $\tilde{P}_2$ on $\tilde{V}_2$ is a permutation on $V$.
For finding such a function $\tilde{P}_2$, in the reduced MIP only the rows corresponding to $\tilde{V}_2$ and the columns corresponding to $V\setminus\Ima \tilde{P}_1$ have to be considered, where $\Ima\tilde{P}_1$ denotes the image of the function $\tilde{P}$.

In principle, any heuristic calculating an incomplete permutation can be chosen for the pre-processing step.
The heuristic we chose is based on the following simple observation: 
If a transformation $P$ maps node $x$ to $x'$, then it can map any node $z$ adjacent to $x$ only to those nodes $z'$ adjacent to $x'$ for which the edge $\{x,z\}$ has the same weight as the edge $\{x', z'\}$.
That way, when committing to an image $\tilde{P}(x)$ of node $x$, we not only gain information about $x$ itself but also about the remaining possible images of all other nodes. The general concept is known as \textit{constraint propagation} \cite{russell_artificial_2021}.
A pseudocode description of this algorithm is given in \Cref{sec:Algorithms}, \Cref{alg:Transformation_Finder}.

We now describe the algorithm for actually finding the invariance transformations in  more detail and refer to \Cref{sec:Algorithms}, \Cref{alg:Full_Algorithm}, for a pseudocode description.
Assume that we are given the set of nodes $V=\{1,\dots,n\}$ and let $L$ denote a set of bins, which form equivalence classes of edges. 
They are used to decide when two edge weights are considered to be identical, which is the case if and only if they belong to the same bin. 
We discuss the binning step in more detail later.

When trying to find an automorphism $P$, the algorithm maintains a set~$\tilde{P}_x$ of possible images for each node $x\in V$.
Initially, when every node could potentially be mapped to any other, we have $\tilde{P}_x = V$ for all $x \in V$. 
The algorithm then tries to refine these sets until for each node, this set has either size $0$ or $1$. 
Ideally, one would like to have $|\tilde{P}_x|=1$ for all $x\in V$, with the intersection of each pair of sets being empty.
This then implies that each node has a unique matching partner, hence $\tilde{P}$ is a \enquote{proper} permutation.
However, as we deal with imperfect data, we accept that a certain \enquote{allowed percentage} of the nodes can not be mapped, which is encoded by the corresponding set $\tilde{P}_x$ being empty.
This percentage is called \emph{fault tolerance}.
Thus, if too many sets $\tilde{P}_x$ are empty at some point, we discard the currently constructed incomplete permutation~$\tilde{P}$.

The algorithm now works as follows. 
First, it checks whether the current incomplete permutation can be discarded.
If this is not the case, it checks whether at least one node can still be potentially mapped to multiple other nodes, branches, and tests all these possibilities. 
To locally minimize the number of times the algorithm branches, it chooses the index $x$ for which $\tilde{P}_x$ has the least number of elements larger than one. 

In each branching step trying to map $x$ to some potential target $x'\in\tilde{P}_x$, the algorithm now applies the core idea to reduce the sets $\tilde{P}_1,\dots,\tilde{P}_n$ even further.
For all bins $\ell\in L$, it calculates the set $S_\ell$ of all nodes $z'$ such that the edge $\{x',z'\}$ is in bin~$\ell$.
Now, for all edges $\{x, z\}$ that are also in bin $\ell$, the node~$z$ can only be mapped to elements in $S_\ell$, so $\tilde{P}_z$ is replaced by its intersection with $S_\ell$. 
This filtering procedure is applied recursively to further narrow down all remaining $\tilde{P}_x$ that contain more than one element until any of the exit conditions above is met.
Depending on which one it was, the incomplete permutation is either discarded or stored for potential later use in the reduced MIP. 

\section{Numerical Experiments} \label{sec:Numerical}

To verify our concepts, we implemented a prototype of the algorithmic framework presented in \Cref{sec:Method}.
We set up the integer program \Cref{equ:Exact_Integer_Program} using the Python package Pyomo \cite{Bynum:2021,Hart:2011} and solved it using the academic solver SCIP \cite{Gamrath:2020a}, Version 7.0.3.
The pre-processing was implemented explicitly and the handling of permutations was mainly done using SymPy \cite{10.7717/peerj-cs.103} and NumPy~\cite{harris2020array}.
The implementation that has been used for the study presented here is available on github \cite{pyaga_github}.

\subsection{Synthetic Data} \label{sec:GeneralExperimentalSetup}

Our experiments were executed on synthetic data, all of which consists of a variety of two-dimensional finite surfaces with binary or colored pixels.
The opposing sides of these surfaces are connected with each other, so the surfaces are embedded on a torus.
For simplicity, we depict these surfaces as two-dimensional Cartesian meshes.

\newcommand{\letter}[4][0]{%
\begin{tikzpicture}[ampersand replacement=\&]
\begin{scope}[local bounding box=a]
\matrix [
         matrix of nodes, 
         nodes={draw=black!30, fill=white,scale=#2,transform shape}, 
         nodes in empty cells,
         column sep=-\pgflinewidth,
         row sep=-\pgflinewidth]{#4};
   \ifthenelse{#1=1}{\node [fit=(a),inner sep=-3.55pt,draw] {};
    }{}
\end{scope}
\end{tikzpicture}%
}%

\newcommand{\f}{|[fill,black]|}

\newcommand{\letterEmpty}[1][0]{%
\letter[#1]{0.5}{A}{%
\&  \&  \& \node(T){}; \&  \&  \&\\
\&  \&  \&  \&  \&  \&\\
\&  \&  \&  \&  \&  \&\\
\node(L){};\&  \&  \&  \&  \&  \&\node(R){};\\
\&  \&  \&  \&  \&  \&\\
\&  \&  \&  \&  \&  \&\\
\&  \&  \&  \node(B){};\&  \&  \&\\
}
}

\newcommand{\letterA}[1][0]{%
\letter[#1]{0.5}{A}{%
\&  \&  \&  \&  \&  \&\\
\&\f\&\f\&\f\&\f\&  \&\\
\&\f\&  \&  \&\f\&  \&\\
\&\f\&  \&  \&\f\&  \&\\
\&\f\&\f\&\f\&\f\&  \&\\
\&\f\&  \&  \&\f\&  \&\\
\&  \&  \&  \&  \&  \&\\
}
}

\newcommand{\letterB}[1][0]{%
\letter[#1]{0.5}{B}{%
\&  \&  \&  \&  \&  \&\\
\&\f\&\f\&\f\&  \&  \&\\
\&\f\&  \&  \&\f\&  \&\\
\&\f\&\f\&\f\&  \&  \&\\
\&\f\&  \&  \&\f\&  \&\\
\&\f\&\f\&\f\&  \&  \&\\
\&  \&  \&  \&  \&  \&\\
}}

\newcommand{\letterC}[1][0]{%
\letter[#1]{0.5}{C}{%
\&  \&  \&  \&  \&  \&\\
\&\f\&\f\&\f\&\f\&  \&\\
\&\f\&  \&  \&  \&  \&\\
\&\f\&  \&  \&  \&  \&\\
\&\f\&  \&  \&  \&  \&\\
\&\f\&\f\&\f\&\f\&  \&\\
\&  \&  \&  \&  \&  \&\\
}}

\newcommand{\letterD}[1][0]{%
\letter[#1]{0.5}{D}{%
\&  \&  \&  \&  \&  \&\\
\&\f\&\f\&\f\&  \&  \&\\
\&\f\&  \&  \&\f\&  \&\\
\&\f\&  \&  \&\f\&  \&\\
\&\f\&  \&  \&\f\&  \&\\
\&\f\&\f\&\f\&  \&  \&\\
\&  \&  \&  \&  \&  \&\\
}}

\newcommand{\letterE}[1][0]{%
\letter[#1]{0.5}{E}{%
\&  \&  \&  \&  \&  \&\\
\&\f\&\f\&\f\&\f\&  \&\\
\&\f\&  \&  \&  \&  \&\\
\&\f\&\f\&\f\&  \&  \&\\
\&\f\&  \&  \&  \&  \&\\
\&\f\&\f\&\f\&\f\&  \&\\
\&  \&  \&  \&  \&  \&\\
}}

\newcommand{\letterF}[1][0]{%
\letter[#1]{0.5}{F}{%
\&  \&  \&  \&  \&  \&\\
\&\f\&\f\&\f\&\f\&  \&\\
\&\f\&  \&  \&  \&  \&\\
\&\f\&\f\&\f\&  \&  \&\\
\&\f\&  \&  \&  \&  \&\\
\&\f\&  \&  \&  \&  \&\\
\&  \&  \&  \&  \&  \&\\
}}

\newcommand{\letterG}[1][0]{%
\letter[#1]{0.5}{G}{%
\&  \&  \&  \&  \&  \&\\
\&  \&\f\&\f\&\f\&  \&\\
\&\f\&  \&  \&  \&  \&\\
\&\f\&  \&\f\&\f\&  \&\\
\&\f\&  \&  \&\f\&  \&\\
\&  \&\f\&\f\&\f\&  \&\\
\&  \&  \&  \&  \&  \&\\
}}

\newcommand{\letterH}[1][0]{%
\letter[#1]{0.5}{H}{%
\&  \&  \&  \&  \&  \&\\
\&\f\&  \&  \&\f\&  \&\\
\&\f\&  \&  \&\f\&  \&\\
\&\f\&\f\&\f\&\f\&  \&\\
\&\f\&  \&  \&\f\&  \&\\
\&\f\&  \&  \&\f\&  \&\\
\&  \&  \&  \&  \&  \&\\
}}

\newcommand{\letterI}[1][0]{%
\letter[#1]{0.5}{I}{%
\&  \&  \&  \&  \&  \&\\
\&\f\&  \&  \&  \&  \&\\
\&\f\&  \&  \&  \&  \&\\
\&\f\&  \&  \&  \&  \&\\
\&\f\&  \&  \&  \&  \&\\
\&\f\&  \&  \&  \&  \&\\
\&  \&  \&  \&  \&  \&\\
}}

\newcommand{\letterJ}[1][0]{%
\letter[#1]{0.5}{J}{%
\&  \&  \&  \&  \&  \&\\
\&  \&\f\&\f\&\f\&  \&\\
\&  \&  \&  \&\f\&  \&\\
\&  \&  \&  \&\f\&  \&\\
\&\f\&  \&  \&\f\&  \&\\
\&  \&\f\&\f\&  \&  \&\\
\&  \&  \&  \&  \&  \&\\
}}

\newcommand{\letterK}[1][0]{%
\letter[#1]{0.5}{K}{%
\&  \&  \&  \&  \&  \&\\
\&\f\&  \&  \&\f\&  \&\\
\&\f\&  \&\f\&  \&  \&\\
\&\f\&\f\&  \&  \&  \&\\
\&\f\&  \&\f\&  \&  \&\\
\&\f\&  \&  \&\f\&  \&\\
\&  \&  \&  \&  \&  \&\\
}}

\newcommand{\letterL}[1][0]{%
\letter[#1]{0.5}{L}{%
\&  \&  \&  \&  \&  \&\\
\&\f\&  \&  \&  \&  \&\\
\&\f\&  \&  \&  \&  \&\\
\&\f\&  \&  \&  \&  \&\\
\&\f\&  \&  \&  \&  \&\\
\&\f\&\f\&\f\&\f\&  \&\\
\&  \&  \&  \&  \&  \&\\
}}

\newcommand{\letterM}[1][0]{%
\letter[#1]{0.5}{M}{%
\&  \&  \&  \&  \&  \&\\
\&\f\&  \&  \&  \&\f\&\\
\&\f\&\f\&  \&\f\&\f\&\\
\&\f\&  \&\f\&  \&\f\&\\
\&\f\&  \&  \&  \&\f\&\\
\&\f\&  \&  \&  \&\f\&\\
\&  \&  \&  \&  \&  \&\\
}}

\newcommand{\letterN}[1][0]{%
\letter[#1]{0.5}{N}{%
\&  \&  \&  \&  \&  \&\\
\&\f\&  \&  \&  \&\f\&\\
\&\f\&\f\&  \&  \&\f\&\\
\&\f\&  \&\f\&  \&\f\&\\
\&\f\&  \&  \&\f\&\f\&\\
\&\f\&  \&  \&  \&\f\&\\
\&  \&  \&  \&  \&  \&\\
}}

\newcommand{\letterO}[1][0]{%
\letter[#1]{0.5}{O}{%
\&  \&  \&  \&  \&  \&\\
\&  \&\f\&\f\&\f\&  \&\\
\&\f\&  \&  \&  \&\f\&\\
\&\f\&  \&  \&  \&\f\&\\
\&\f\&  \&  \&  \&\f\&\\
\&  \&\f\&\f\&\f\&  \&\\
\&  \&  \&  \&  \&  \&\\
}}

\newcommand{\letterP}[1][0]{%
\letter[#1]{0.5}{P}{%
\&  \&  \&  \&  \&  \&\\
\&\f\&\f\&\f\&  \&  \&\\
\&\f\&  \&  \&\f\&  \&\\
\&\f\&\f\&\f\&  \&  \&\\
\&\f\&  \&  \&  \&  \&\\
\&\f\&  \&  \&  \&  \&\\
\&  \&  \&  \&  \&  \&\\
}}

\newcommand{\letterQ}[1][0]{%
\letter[#1]{0.5}{Q}{%
\&  \&  \&  \&  \&  \&\\
\&  \&\f\&\f\&\f\&  \&\\
\&\f\&  \&  \&  \&\f\&\\
\&\f\&  \&\f \&  \&\f\&\\
\&\f\&  \&  \&\f\&  \&\\
\&  \&\f\&\f\&  \&\f\&\\
\&  \&  \&  \&  \&  \&\\
}}

\newcommand{\letterR}[1][0]{%
\letter[#1]{0.5}{R}{%
\&  \&  \&  \&  \&  \&\\
\&\f\&\f\&\f\&  \&  \&\\
\&\f\&  \&  \&\f\&  \&\\
\&\f\&\f\&\f\&  \&  \&\\
\&\f\&  \&\f\&  \&  \&\\
\&\f\&  \&  \&\f\&  \&\\
\&  \&  \&  \&  \&  \&\\
}}

\newcommand{\letterS}[1][0]{%
\letter[#1]{0.5}{S}{%
\&  \&  \&  \&  \&  \&\\
\&\f\&\f\&\f\&\f\&  \&\\
\&\f\&  \&  \&  \&  \&\\
\&\f\&\f\&\f\&\f\&  \&\\
\&  \&  \&  \&\f\&  \&\\
\&\f\&\f\&\f\&\f\&  \&\\
\&  \&  \&  \&  \&  \&\\
}}

\newcommand{\letterT}[1][0]{%
\letter[#1]{0.5}{T}{%
\&  \&  \&  \&  \&  \&\\
\&\f\&\f\&\f\&\f\&\f\&\\
\&  \&  \&\f\&  \&  \&\\
\&  \&  \&\f\&  \&  \&\\
\&  \&  \&\f\&  \&  \&\\
\&  \&  \&\f\&  \&  \&\\
\&  \&  \&  \&  \&  \&\\
}}

\newcommand{\letterU}[1][0]{%
\letter[#1]{0.5}{U}{%
\&  \&  \&  \&  \&  \&\\
\&\f\&  \&  \&  \&\f\&\\
\&\f\&  \&  \&  \&\f\&\\
\&\f\&  \&  \&  \&\f\&\\
\&\f\&  \&  \&  \&\f\&\\
\&  \&\f\&\f\&\f\&  \&\\
\&  \&  \&  \&  \&  \&\\
}}

\newcommand{\letterV}[1][0]{%
\letter[#1]{0.5}{V}{%
\&  \&  \&  \&  \&  \&\\
\&\f\&  \&  \&  \&\f\&\\
\&\f\&  \&  \&  \&\f\&\\
\&\f\&  \&  \&  \&\f\&\\
\&  \&\f\&  \&\f\&  \&\\
\&  \&  \&\f\&  \&  \&\\
\&  \&  \&  \&  \&  \&\\
}}

\newcommand{\letterW}[1][0]{%
\letter[#1]{0.5}{W}{%
\&  \&  \&  \&  \&  \&\\
\&\f\&  \&  \&  \&\f\&\\
\&\f\&  \&  \&  \&\f\&\\
\&\f\&  \&\f\&  \&\f\&\\
\&\f\&\f\&  \&\f\&\f\&\\
\&\f\&  \&  \&  \&\f\&\\
\&  \&  \&  \&  \&  \&\\
}}

\newcommand{\letterX}[1][0]{%
\letter[#1]{0.5}{X}{%
\&  \&  \&  \&  \&  \&\\
\&\f\&  \&  \&  \&\f\&\\
\&  \&\f\&  \&\f\&  \&\\
\&  \&  \&\f\&  \&  \&\\
\&  \&\f\&  \&\f\&  \&\\
\&\f\&  \&  \&  \&\f\&\\
\&  \&  \&  \&  \&  \&\\
}}

\newcommand{\letterY}[1][0]{%
\letter[#1]{0.5}{Y}{%
\&  \&  \&  \&  \&  \&\\
\&\f\&  \&  \&  \&\f\&\\
\&  \&\f\&  \&\f\&  \&\\
\&  \&  \&\f\&  \&  \&\\
\&  \&\f\&  \&  \&  \&\\
\&\f\&  \&  \&  \&  \&\\
\&  \&  \&  \&  \&  \&\\
}}

\newcommand{\letterZ}[1][0]{%
\letter[#1]{0.5}{Z}{%
\&  \&  \&  \&  \&  \&\\
\&\f\&\f\&\f\&\f\&  \&\\
\&  \&  \&  \&\f\&  \&\\
\&  \&\f\&\f\&  \&  \&\\
\&\f\&  \&  \&  \&  \&\\
\&\f\&\f\&\f\&\f\&  \&\\
\&  \&  \&  \&  \&  \&\\
}}

\newcommand{\alphabet}{%
  \letterA%
  \letterB%
  \letterC%
  \letterD%
  \letterE%
  \letterF%
  \letterG%
  \letterH%
  \letterI%
  \letterJ%
  \letterK%
  \letterL%
  \letterM%
  
  \letterN%
  \letterO%
  \letterP%
  \letterQ%
  \letterR%
  \letterS%
  \letterT%
  \letterU%
  \letterV%
  \letterW%
  \letterX%
  \letterY%
  \letterZ%
}

\newcommand{\worldAI}[1][0]{%
\letter[#1]{0.44}{AI}{%
  \&  \&  \&  \&  \&  \&  \&  \&  \&  \&  \&  \&  \&  \&  \&  \&  \&  \&  \&  \\
  \&  \&  \&  \&  \&  \&  \&  \&  \&  \&  \&  \&  \&  \&  \&  \&  \&  \&  \&  \\
  \&  \&  \&  \&  \&  \&  \&  \&  \&  \&  \&  \&  \&  \&  \&  \&  \&  \&  \&  \\
  \&  \&  \&  \&  \&  \&  \&\f\&\f\&\f\&\f\&  \&\f\&  \&  \&  \&  \&  \&  \&  \\
  \&  \&  \&  \&  \&  \&  \&\f\&  \&  \&\f\&  \&\f\&  \&  \&  \&  \&  \&  \&  \\
  \&  \&  \&  \&  \&  \&  \&\f\&  \&  \&\f\&  \&\f\&  \&  \&  \&  \&  \&  \&  \\
  \&  \&  \&  \&  \&  \&  \&\f\&\f\&\f\&\f\&  \&\f\&  \&  \&  \&  \&  \&  \&  \\
  \&  \&  \&  \&  \&  \&  \&\f\&  \&  \&\f\&  \&\f\&  \&  \&  \&  \&  \&  \&  \\
  \&  \&  \&  \&  \&  \&  \&  \&  \&  \&  \&  \&  \&  \&  \&  \&  \&  \&  \&  \\
  \&  \&  \&  \&  \&  \&  \&  \&  \&  \&  \&  \&  \&  \&  \&  \&  \&  \&  \&  \\
}}

\newcommand{\worldLR}[1][0]{%
\letter[#1]{0.44}{LR}{%
  \&  \&  \&  \&  \&  \&  \&  \&  \&  \&  \&  \&  \&  \&  \&  \&  \&  \&  \&  \\
  \&  \&  \&  \&  \&  \&  \&  \&  \&  \&  \&  \&  \&  \&  \&  \&  \&  \&  \&  \\
  \&  \&  \&  \&  \&  \&  \&  \&  \&  \&  \&  \&\f\&  \&  \&  \&  \&\f\&\f\&\f\\
\f\&  \&  \&  \&  \&  \&  \&  \&  \&  \&  \&  \&\f\&  \&  \&  \&  \&\f\&  \&  \\
  \&  \&  \&  \&  \&  \&  \&  \&  \&  \&  \&  \&\f\&  \&  \&  \&  \&\f\&\f\&\f\\
  \&  \&  \&  \&  \&  \&  \&  \&  \&  \&  \&  \&\f\&  \&  \&  \&  \&\f\&  \&\f\\
\f\&  \&  \&  \&  \&  \&  \&  \&  \&  \&  \&  \&\f\&\f\&\f\&\f\&  \&\f\&  \&  \\
  \&  \&  \&  \&  \&  \&  \&  \&  \&  \&  \&  \&  \&  \&  \&  \&  \&  \&  \&  \\
  \&  \&  \&  \&  \&  \&  \&  \&  \&  \&  \&  \&  \&  \&  \&  \&  \&  \&  \&  \\
  \&  \&  \&  \&  \&  \&  \&  \&  \&  \&  \&  \&  \&  \&  \&  \&  \&  \&  \&  \\
}}

\newcommand{\worldGG}[1][0]{%
\letter[#1]{0.44}{GG}{%
  \&  \&  \&  \&  \&  \&  \&  \&  \&  \&  \&  \&  \&  \&  \&  \&  \&  \&  \&  \\
  \&  \&  \&  \&  \&  \&  \&  \&  \&  \&  \&  \&  \&  \&  \&  \&  \&  \&  \&  \\
  \&  \&  \&  \&  \&  \&  \&  \&  \&  \&  \&  \&  \&  \&  \&  \&  \&  \&  \&  \\
  \&  \&  \&  \&  \&  \&  \&  \&\f\&\f\&\f\&  \&  \&\f\&\f\&\f\&  \&  \&  \&  \\
  \&  \&  \&  \&  \&  \&  \&\f\&  \&  \&  \&  \&\f\&  \&  \&  \&  \&  \&  \&  \\
  \&  \&  \&  \&  \&  \&  \&\f\&  \&\f\&\f\&  \&\f\&  \&\f\&\f\&  \&  \&  \&  \\
  \&  \&  \&  \&  \&  \&  \&\f\&  \&  \&\f\&  \&\f\&  \&  \&\f\&  \&  \&  \&  \\
  \&  \&  \&  \&  \&  \&  \&  \&\f\&\f\&\f\&  \&  \&\f\&\f\&\f\&  \&  \&  \&  \\
  \&  \&  \&  \&  \&  \&  \&  \&  \&  \&  \&  \&  \&  \&  \&  \&  \&  \&  \&  \\
  \&  \&  \&  \&  \&  \&  \&  \&  \&  \&  \&  \&  \&  \&  \&  \&  \&  \&  \&  \\
}}

\newcommand{\worldWP}[1][0]{%
\letter[#1]{0.44}{WP}{%
  \&  \&  \&  \&  \&  \&  \&  \&  \&  \&\f\&  \&\f\&  \&\f\&  \&\f\&\f\&\f\&  \\
  \&  \&  \&  \&  \&  \&  \&  \&  \&  \&\f\&\f\&  \&\f\&\f\&  \&\f\&  \&  \&  \\
  \&  \&  \&  \&  \&  \&  \&  \&  \&  \&\f\&  \&  \&  \&\f\&  \&\f\&  \&  \&  \\
  \&  \&  \&  \&  \&  \&  \&  \&  \&  \&  \&  \&  \&  \&  \&  \&  \&  \&  \&  \\
  \&  \&  \&  \&  \&  \&  \&  \&  \&  \&  \&  \&  \&  \&  \&  \&  \&  \&  \&  \\
  \&  \&  \&  \&  \&  \&  \&  \&  \&  \&  \&  \&  \&  \&  \&  \&  \&  \&  \&  \\
  \&  \&  \&  \&  \&  \&  \&  \&  \&  \&  \&  \&  \&  \&  \&  \&  \&  \&  \&  \\
  \&  \&  \&  \&  \&  \&  \&  \&  \&  \&  \&  \&  \&  \&  \&  \&  \&  \&  \&  \\
  \&  \&  \&  \&  \&  \&  \&  \&  \&  \&\f\&  \&  \&  \&\f\&  \&\f\&\f\&\f\&  \\
  \&  \&  \&  \&  \&  \&  \&  \&  \&  \&\f\&  \&  \&  \&\f\&  \&\f\&  \&  \&\f\\
}}

\newcommand{\worldLU}[1][0]{%
\letter[#1]{0.44}{LU}{%
  \&  \&  \&  \&  \&  \&  \&  \&  \&  \&  \&  \&  \&  \&  \&  \&  \&  \&  \&  \\
  \&  \&  \&  \&  \&  \&  \&  \&  \&  \&  \&  \&  \&  \&  \&  \&  \&  \&  \&  \\
  \&  \&  \&  \&  \&  \&  \&  \&  \&  \&  \&  \&  \&  \&  \&  \&  \&  \&  \&  \\
  \&  \&  \&  \&  \&  \&  \&  \&  \&  \&  \&  \&  \&  \&  \&  \&  \&  \&  \&  \\
  \&  \&  \&  \&  \&  \&  \&  \&  \&  \&  \&  \&  \&  \&  \&  \&  \&  \&  \&  \\
  \&\f\&  \&  \&  \&\f\&  \&  \&  \&  \&  \&  \&  \&  \&  \&  \&\f\&  \&  \&  \\
  \&\f\&  \&  \&  \&\f\&  \&  \&  \&  \&  \&  \&  \&  \&  \&  \&\f\&  \&  \&  \\
  \&\f\&  \&  \&  \&\f\&  \&  \&  \&  \&  \&  \&  \&  \&  \&  \&\f\&  \&  \&  \\
  \&\f\&  \&  \&  \&\f\&  \&  \&  \&  \&  \&  \&  \&  \&  \&  \&\f\&  \&  \&  \\
  \&  \&\f\&\f\&\f\&  \&  \&  \&  \&  \&  \&  \&  \&  \&  \&  \&\f\&\f\&\f\&\f\\
}}

\newcommand{\worldCK}[1][0]{%
\letter[#1]{0.44}{CK}{%
  \&  \&\f\&  \&\f\&  \&  \&  \&  \&  \&  \&  \&  \&  \&  \&  \&  \&\f\&  \&  \\
\f\&  \&\f\&  \&  \&\f\&  \&  \&  \&  \&  \&  \&  \&  \&  \&  \&  \&\f\&\f\&\f\\
  \&  \&  \&  \&  \&  \&  \&  \&  \&  \&  \&  \&  \&  \&  \&  \&  \&  \&  \&  \\
  \&  \&  \&  \&  \&  \&  \&  \&  \&  \&  \&  \&  \&  \&  \&  \&  \&  \&  \&  \\
  \&  \&  \&  \&  \&  \&  \&  \&  \&  \&  \&  \&  \&  \&  \&  \&  \&  \&  \&  \\
  \&  \&  \&  \&  \&  \&  \&  \&  \&  \&  \&  \&  \&  \&  \&  \&  \&  \&  \&  \\
  \&  \&  \&  \&  \&  \&  \&  \&  \&  \&  \&  \&  \&  \&  \&  \&  \&  \&  \&  \\
\f\&  \&\f\&  \&  \&\f\&  \&  \&  \&  \&  \&  \&  \&  \&  \&  \&  \&\f\&\f\&\f\\
  \&  \&\f\&  \&\f\&  \&  \&  \&  \&  \&  \&  \&  \&  \&  \&  \&  \&\f\&  \&  \\
  \&  \&\f\&\f\&  \&  \&  \&  \&  \&  \&  \&  \&  \&  \&  \&  \&  \&\f\&  \&  \\
}}

\begin{figure}[htb]
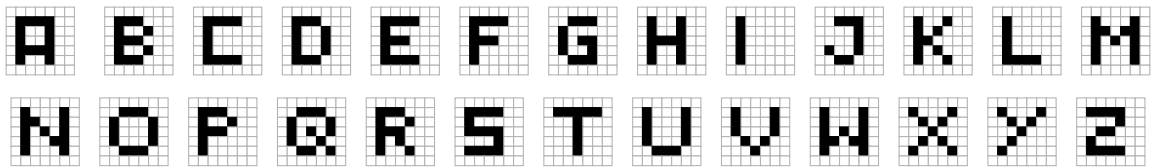

    \centering
    \alphabet
    \caption{Letters composed of binary or colored pixels are used in our synthetic data.}
    \label{fig:alphabet}
\end{figure}

For our experiments, we designed five different sets of synthetic data, also called \emph{worlds}.
A summary, as well as short-hand names for the worlds we used, can be found in \Cref{tab:Setup}.
We consider four different sets of surfaces with $20\times 10, 20\times 20, 15\times 15$ and $13\times 7$ pixels.
In each setting, an individual observation consists of two or three letters placed next to each other with one pixel space between them on the torus.
Our embedding of the letters is shown in \Cref{fig:alphabet}.
In the world T, each letter from \Cref{fig:alphabet} is used.

\begin{figure}[htb]
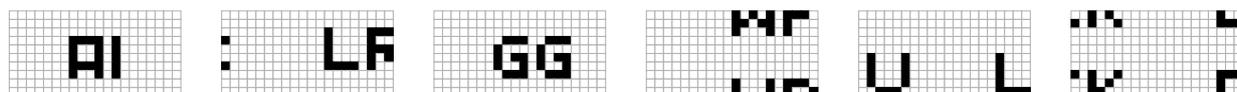

    \centering
    \worldAI\hfill\worldLR\hfill\worldGG\hfill\worldWP\hfill\worldLU\hfill\worldCK
    \caption{Example observations of the test world T. 
    The entities are \enquote{words}, made up of two letters of the English alphabet placed next to each other.}
    \label{fig:examplePictureAI}
\end{figure}

Thus, the full data set consists of all pairs of letters at all possible positions. Some example observations from this world are shown in \Cref{fig:examplePictureAI}.
We also considered two worlds of size $15\times15$, called TR1 and TR2.

In TR1, again all letters are used, while in TR2, we only considered letters that do not exhibit axial symmetry in our representation, meaning only the $11$ letters F, G, J, L, N, P, Q, R, S, Y, Z are used. In both cases, each group of letters can be rotated by $0\degree,90\degree,180\degree$ or $270\degree$.

In TC, the same reduced set of letters is used, but each letter is assigned one of three colors and the size is reduced to $13\times 7$ pixels.
Lastly, we considered TL, a world of size $20\times20$ with three letters, rotations and translations. In addition to the scenario before, individual letters can move vertically as well relative to the word axis, meaning that if the word is rotated by 90°/270°, the individual letter translations are horizontal. The third letter and additional degrees of freedom lead to the number of observations to be drastically higher than for the other worlds. 

The properties of all worlds are summarized in \Cref{tab:Setup}.

\begin{table}[htb]
    \centering
    \begin{tabular}{|c|c|c|c|c|c|c|c|}
        \hline
        & &  & Letters  & \multicolumn{2}{c|}{Letter} & \# of different & Expected \\
        \multirow{2}{*}{Name} & \multirow{2}{*}{Size} & \multirow{2}{*}{Alphabet} & per obser-  & \multicolumn{2}{c|}{transformations} & possible & graph\\
        &  &  & vation  & Type & Count & observations & symmetries\\
        \hline
        T   &$20\times 10$    & full           & 2 & T     & \numprint{200}     &\numprint{135200}       & \numprint{400} \\
        TR1 & $15\times 15$   & full           & 2 & T, R  & \numprint{900}     &\numprint{571950}       & \numprint{900} \\
        TR2 & $15\times 15$   & F, G, \dots, Z & 2 & T, R  & \numprint{900}     &\numprint{104850}       & \numprint{900} \\
        TC  & $13\times 7$    & F, G, \dots, Z & 2 & T, C  & \numprint{546}     & \numprint{99099}       & \numprint{1092}\\
        TL  & $20\times 20$   & full           & 3 & T+, R & \numprint{640000} &\numprint{11248640000}  & \numprint{1600} \\
        \hline
    \end{tabular}
    %
    %
    %
    \caption{Overview over the experiments performed to test the algorithmic framework.
    The letters R, C signify rotational and color change transformations, respectively. T indicates global translations only, while T+ means that horizontal translations are only applicable to all letters simultaneously and vertical translations can be applied to individual letters. Note that some combinations of letters like \enquote{OO} are symmetric under rotation and therefore the number of different possible observations can be smaller than the combinatorial product of the numbers of possible letters and transformations.}
    \label{tab:Setup}
\end{table}

The letter transformations described above, which are used to generate the observations, are related to the invariance transformations, which we hope to recover, in the following way.
If a letter transformation can be expressed as a global permutation of pixels (like global rotations or translations) or global color change then it automatically leads to a symmetry of $\Psi$ and thus an invariance transformation. 
This is not true for every letter transformation though: For example, a vertical shift of an individual letter (as in TL) cannot be written as one permutation matrix acting on $\featspace$, because the position of the letters differ between the observations.
Thus there is no reason to expect an invariance transformation or a symmetry of the concurrence graph to arise from such letter transformations.
Consequently, the number of graph symmetries in TL is much smaller than the number of letter transformations (\cf  \Cref{tab:Setup}).

Conversely, our particular method of finding invariances can lead to false positives.
As explained in \Cref{sec:transformations}, we replace the hard problem of finding symmetries of $\Psi$ by the simpler one of finding symmetries in its projections onto one or two coordinates of $\featspace$ (\ite the concurrence graph), and while a symmetry of $\Psi$ always gives rise to a symmetry of the concurrence graph, the opposite is not always true. 
In particular, given a torus and global translational symmetry on its surface, the marginal distribution does not change under a rotation by $180\degree$ around any one point:
$\Psi_i$ is the same for all $i$ anyway, and $\Psi_{ij}$ only depends on the relative vector from $i$ to $j$. A $180\degree$ rotation maps $(i,j)$ to $(i',j')$, inverting the relative vector. But since $\Psi_{ij} = \Psi_{ji}$ we also have $\Psi_{ij} = \Psi_{i'j'}$, making the $180\degree$ rotation a symmetry of the concurrence graph. 
But if $x\in\featspace$ denotes some observation that does not possess rotational symmetry by itself, then a rotated version $x_r\in\featspace$ of this feature does not necessarily fulfill $\dist(x_r) = \dist(x)$, meaning that \dist itself does not possess this rotational symmetry.
Consequently, the $180\degree$ rotation is a false positive for distributions $\Psi$ which are invariant under translation but not under rotation and therefore in the worlds T and TC we expect our algorithm to find twice as many graph symmetries as there are letter transformations, \cf \Cref{tab:Setup}.

\subsection{Specific Test Cases and Parameters}

\begin{figure}[htb]
    \centering
    \includegraphics[width=0.99\textwidth]{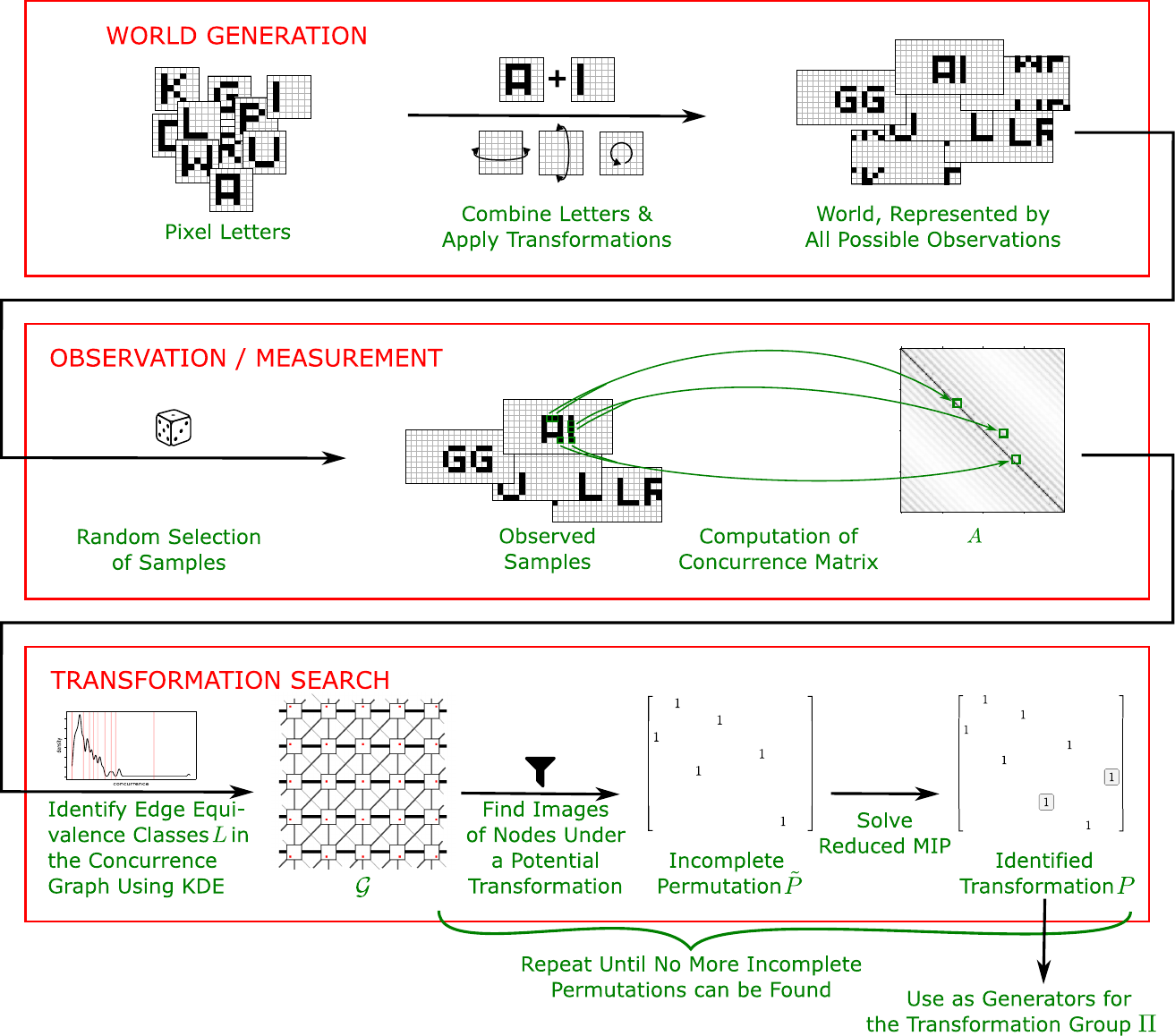}
    \caption{Workflow of the numerical experiments}
    \label{fig:workflow}
\end{figure}

\Cref{fig:workflow} summarizes the overall workflow of our numerical studies.
For all experiments we first generated synthetic data sets given by observations of the worlds described in \Cref{tab:Setup}. These data sets contained different percentages of all possible observations in order to be able to study how effectively the method generalizes from few observations. To perform a broad analysis, the used ratios ranging from $5\%$ up to $100\%$ in increments of $5\%$ or $10\%$ for all worlds but TL, adjusting the grid resolution in a trade-off of exploration of possible performance variance and runtime. For the world TL, our goal was to test the limits of our method, we considered a range from $0.004\%$ to $5\%$. For each data set, we then calculated the concurrence graph and applied \Cref{alg:Full_Algorithm} to find the invariance transformations.



The algorithm depends on several hyperparameters, namely the error value limit, as well as the bandwidth and fault tolerance. The latter two parameters define the resillience of the algorithm w.r.t. perturbances in the edge weights of the adjacency matrix by defining the classes of weights which are considered the same (i.e. binning) as well as the tolerance to possible misclassifications, while the error value limit is explained in more detail below.
We chose bins of variable sizes using a kernel density estimation, a non-parametric method to approximate the underlying density function from a given set of samples~\cite{KDE_Tutorial}.
We used a Gaussian kernel and performed each experiment with a fixed bandwidth.

From the edge weight density estimation, we defined the bins to be the halfway-points between two consecutive maxima.
Additionally, we defined a-priori that edges with zero weight are grouped separately.

For our experiments, it is sufficient to find a suitable set of generators of the respective symmetry group.
For the worlds with only translational and rotational invariances, one such set can consist, for example, of three permutations.
These represent horizontal, vertical and rotational transformations, respectively the $180\degree$ rotation.
In the experiment involving colored letters, this set has to be extended by two additional generators.
They realize the color changes via an exchange of either all three or just two colors.

In the presented framework, solving the MIP is by far the hardest part and it is desirable to minimize the number of times it has to be solved.
To this end, we employed the following strategy:
We maintain a set $\mathcal{G}$ of generators for the symmetry group as well as the symmetry group \group itself which we both initialize as empty sets.
We then calculate the set of all incomplete permutations using the pre-processing step described in \Cref{sec:SolveMIPInPractice} resp. \Cref{alg:Transformation_Finder} in \Cref{sec:Algorithms} and iterate over all members of this set:
for each incomplete permutation $\tilde{P}$ which is then only defined on a subset $\tilde{V}$ of $V$, we check if there is already a permutation $P\in\group$ that acts exactly as the currently considered incomplete permutation $\tilde{P}$ on $\tilde{V}$.
If this is the case, then there is no need to complete~$\tilde{P}$ and we continue with the next incomplete permutation.
Otherwise, we complete~$\tilde{P}$ using the reduced MIP and obtain a full permutation $P:V\to V$.
We then verify if this permutation is a sufficiently good approximate automorphism by calculating the devation value $||AP-PA||_\infty$ and checking if it is smaller than a value which we will call the error value limit.
If this is the case, then $P$ is added to the set of generators, i.e., we set $\mathcal{G}\coloneqq\mathcal{G}\cup\{P\}$.
Consequently, we calculate all permutations that can now be created using this larger set of generators, include these in \group and iterate.
This exploitation of the group structure of the permutations enables us to find a significant number of new permutations by solving the MIP only a few times.

A pseudocode summary of this algorithm can be found in \Cref{sec:Algorithms}, \Cref{alg:Full_Algorithm}.

\subsection{Results and Discussion} \label{sec:Results}

We found that pre-processing enabled us to find transformations in cases in which solving the MIP (\ref{equ:Exact_Integer_Program}) by itself was computationally intractable because of the huge number of feasible solutions.
We attribute this to the structure of the worlds from which we sampled our data, since only very specific transformations are present in our experiments.
For translations, fixing $\tilde{P}(i)$ uniquely determines all $\tilde{P}(j)$ for $j\neq i$.
For rotations, this is not as strict as our observations are embedded on a torus.
However, this does not trivialize the problem of finding these invariances as the matrix $A$ and the concurrence graph itself generally does not reflect any information about the structure of the world.
Nevertheless, a commitment to a specific mapping reduced the possible targets of the other nodes considerably, since the structure of the worlds allowed effective propagation of the consequent constraints. 
In particular, committing to incompatible maps of only two nodes is often enough for the algorithm to return early and to try and find another way to construct the permutation.


    \begin{figure}
        \centering
        \input{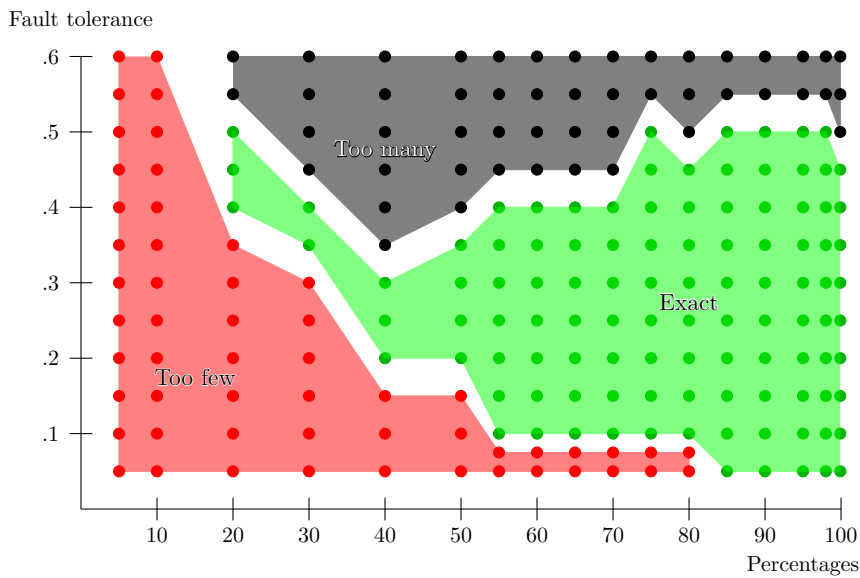}
        \caption{Results of the experiments performed in the $15\times15$-sized world with translational and rotational invariance (TR1).
        We used a KDE bandwidth of $6.4633\cdot 10^{-5}$ and an error value limit of $0.01$ for all experiments.
        Different colors of the symbols denote that too few/the correct number of/too many transformations were calculated.
        The colored regions are included to improve the presentation of the results.}
        \label{fig:1515FullExperiments}
    \end{figure}

For the worlds T, TR1, TR2 and TC with few possible observations, we are able to reliably find the correct set of transformations, even if only $20\%$ or $30\%$ of the observations are available. For the TL world, we succeeded in finding the transformations when \numprint{499982} unique observations, i.e.,  $0.004\%$ of all possible observations of this world were available.
Especially in this range of percentages, the choice of a suitable bandwidth and fault tolerance parameter are important and not trivial, as can be with the example of the experiments conducted on TR1, see \Cref{fig:1515FullExperiments}.
While specifically the fault tolerance has little to no influence on finding the transformations if the observed statistics are sufficiently good, setting it too low resp. too high can result in not all resp. too many transformation generators being calculated.
Here, used bandwidth was $6.4633\cdot 10^{-5}$; the values chosen for the other worlds were in the same order of magnitude, yet specific to the worlds themselves and similarly extensive experiments on the worlds T, TR2 and TC allowed for the same conclusions.

It is not surprising that both a low fault tolerance and a limited number of observations result in too few found transformations.
Interestingly, our algorithm often identifies additional permutations as transformations for intermediate percentages and a high fault tolerance.
This happens as the statistics are only slightly disturbed in this range of percentages, resulting in permutations that are \enquote{nearly} transformations having an error value that is below the error value limit.
An example is a mirror operation along either the horizontal or the vertical axis as most of our letters either exhibit such a mirror symmetry directly or are only a few pixels off, see \Cref{fig:alphabet}.
For this reason, we decided to exclude letters exhibiting axial symmetry in the two experiments TR2 and TC.

With this reduced set of observations, the algorithm then does not find too many transformations compared to the case where all letters are used for creating the observations.
However, finding the expected transformations is then harder, as the total number of unique observations is smaller and statistics deteriorate faster.

Values for the bandwidth and the error value limit which are too high might cause permutations to be accepted which do not correspond to invariance transformations as defined in \Cref{sec:transformations}, while values which are too low might preclude (some) invariance transformations from being found.

\section{Conclusion} \label{sec:Conclusion}
In this paper, we numerically confirmed and illustrated that invariance transformations can be learned in an unsupervised way from the statistics of input data and without prior assumptions on the perceptual modality.
By translating the features that are present in the data and their concurrences into a graph, these transformations can be considered as approximate graph automorphisms.
We implemented a prototype that uses a pre-solving heuristic and a mixed integer program in order to verify that these invariance transformations can be found in practice. The concept of directly defining approximate graph algorithms via solutions of a mixed integer program is a novel approach, and the authors are not aware of any publications investigating approximate automorphisms in this way.

While the presented concept for inferring invariance transformations is inspired by the field of neuroscience, the algorithmic framework is certainly very different from how biological brains work. However, the presented theory and algorithm establish that invariance transformations can in principle be encoded in the structure of the graph of recurrent synaptic connections.

\clearpage

\appendix

\section{Algorithms} \label{sec:Algorithms}

\begin{algorithm}[!htb]
  \SetAlgoLined\DontPrintSemicolon\LinesNumbered
  \SetKwFunction{calcPerms}{calculate\_perms}
  \SetKwFunction{calcIncompletePerms}{calculate\_incomplete\_permutations}
  \SetKwProg{myalg}{Algorithm}{}{}
  \SetKwProg{myproc}{Subroutine}{}{}

  \KwIn{concurrence graph $\congraph$, bins $L$}
  \KwOut{set $\mathtt{found\_trafos}$ containing all possible transformations}
  
  \myalg{\calcIncompletePerms{$\congraph,L$}}{
  $\mathtt{found\_perms}\coloneqq\emptyset$\;
  $\tilde{P}_i\coloneqq V$ for all $i\in V$\tcp*{$\tilde{P}_z:$ set of potential targets for $z$}
    \calcPerms{$\tilde{P}_1,\dots,\tilde{P}_n$}\;
    \KwRet $\mathtt{found\_perms}$\;
  }
  
    \setcounter{AlgoLine}{0}
    \myproc{\calcPerms{$\tilde{P}_1,\dots,\tilde{P}_n$}}{
          \If{$|\tilde{P}_i| = 0$ for more than an \enquote{fault tolerance (percentage)} of nodes $i \in V$}{
          \KwRet \tcp*{incompl. permutation is discarded}
        }
            \ElseIf{there are nodes $z\in V$ with $|\tilde{P}_z|>1$}{
              $x \leftarrow $ node $z$ minimizing $|\tilde{P}_z|$ s.t. $|\tilde{P}_z|>1$ \tcp*{node with few pot. targets}
            \ForEach{$x'\in\tilde{P}_x$}{
                $\widehat{P}_z\leftarrow\tilde{P}_z$ for all $z\in V$ \tcp*{copy potential targets}
                \ForEach{bin $\ell\in L$}{
                    $S_{\ell}\leftarrow$ set of nodes $z'$ adjacent to $x'$ s.t. edge $\{x',z'\}$ is in bin $\ell$\;
                    \ForEach{edge $\{x,z\}$ in bin $\ell$}{
                        $\widehat{P}_z \leftarrow \tilde{P}_z \cap S_{\ell}$\;
                    }
                }
                \calcPerms{$\widehat{P}_1,\dots,\widehat{P}_n$}\;
            }
        }
         \Else{
          Add $\tilde{P}$ to $\mathtt{found\_perms}$\;
        }
    }
  \caption{The pre-solving routine for calculating incomplete permutations.}
  \label{alg:Transformation_Finder}
\end{algorithm}

\begin{algorithm}[!htb]
  \SetAlgoLined\DontPrintSemicolon
  \SetKwFunction{calcIncompletePerms}{calculate\_incomplete\_permutations}
  \KwIn{concurrence graph $\congraph$, bins $L$}
  \KwOut{set of generators}
  
    \nl $\group\leftarrow\emptyset$ \tcp*{initialize empty group of found permutations}
    \nl $\mathtt{found\_perms}\leftarrow $\calcIncompletePerms{$\congraph, L$}\;
    \nl \ForEach{$\tilde{P}\in \mathtt{found\_perms}$}{
        \nl \If{$\tilde{P}$ does not match any permutation in \group}{
            \nl Calculate complete permutation $P$ from $\tilde{P}$ using the reduced MIP\;
            \nl \If{maximum entry of $|AP-PA|$ does not exceed error value limit \errorlimit}{
                \nl Update \group with $P$\;
        }
    }
  }
  \nl \KwRet generators of $\group$\;

  \caption{The algorithm for calculating the invariance transformations.} \label{alg:Full_Algorithm}
\end{algorithm}

\clearpage

\printbibliography
\ifcsname answerYes\endcsname%
\section*{Checklist}
\begin{enumerate}

\item For all authors...
\begin{enumerate}
  \item Do the main claims made in the abstract and introduction accurately reflect the paper's contributions and scope?
    \answerYes{}
  \item Did you describe the limitations of your work?
    \answerYes{See \Cref{sec:Results}.}
  \item Did you discuss any potential negative societal impacts of your work?
    \answerNA{Being a purely foundational work for analyzing data, our algorithm cannot be used to generate malicious data.
    All data used for analyzing our approach is artificial and was created by the authors solely for the presented analysis.}
  \item Have you read the ethics review guidelines and ensured that your paper conforms to them?
    \answerYes{}
\end{enumerate}

\item If you are including theoretical results...
\begin{enumerate}
  \item Did you state the full set of assumptions of all theoretical results?
    \answerNA{This work is mainly experimental and does not contain original theoretical results.
    All theoretical aspects discussed in \Cref{sec:transformations} were not developed in this work but in \cite{Linde:2021} (incl. full proofs resp. arguments) and are referenced in the main text.}
  \item Did you include complete proofs of all theoretical results?
    \answerNA{See above.}
\end{enumerate}

\item If you ran experiments...
\begin{enumerate}
  \item Did you include the code, data, and instructions needed to reproduce the main experimental results (either in the supplemental material or as a URL)?
    \answerYes{The source code is provided as supplementary material.
    In addition, a publication as a python package is under preparation and will be done after the double-blind peer-review process.}
  \item Did you specify all the training details (e.g., data splits, hyperparameters, how they were chosen)?
    \answerYes{We stated all non-default parameters in \Cref{tab:Setup}.}
  \item Did you report error bars (e.g., with respect to the random seed after running experiments multiple times)?
    \answerNA{As our work constitutes a \enquote{proof of concept} result for a novel algorithmic approach, we focused on a qualitative investigation of our approach and did not include a detailed sensitivity analysis.
    This could be performed in a future work.}
  \item Did you include the total amount of compute and the type of resources used (e.g., type of GPUs, internal cluster, or cloud provider)?
    \answerYes{See beginning of \Cref{sec:Numerical}.}
\end{enumerate}

\item If you are using existing assets (e.g., code, data, models) or curating/releasing new assets...
\begin{enumerate}
  \item If your work uses existing assets, did you cite the creators?
    \answerYes{See beginning of \Cref{sec:Numerical} for code.
    All data has been created by the authors.}
  \item Did you mention the license of the assets?
    \answerNA{All used software is available for academic use, see above, and data has been created by the authors.}
  \item Did you include any new assets either in the supplemental material or as a URL?
    \answerYes{Data and code are provided as supplementary material and will be part of the package that is still to be published.}
  \item Did you discuss whether and how consent was obtained from people whose data you're using/curating?
    \answerNA{All data has been created by the authors and are purely artificial.}
  \item Did you discuss whether the data you are using/curating contains personally identifiable information or offensive content?
    \answerNA{See above.}
\end{enumerate}

\item If you used crowdsourcing or conducted research with human subjects...
\answerNA{All of the following points are not applicable as we neither used crowdsourcing nor conducted research with human subjects.}
\begin{enumerate}
  \item Did you include the full text of instructions given to participants and screenshots, if applicable?
  \item Did you describe any potential participant risks, with links to Institutional Review Board (IRB) approvals, if applicable?
  \item Did you include the estimated hourly wage paid to participants and the total amount spent on participant compensation?
\end{enumerate}

\end{enumerate}
\fi%

\end{document}